\documentclass[conference]{IEEEtran}
\IEEEoverridecommandlockouts

\usepackage{cite}
\usepackage{amsmath,amssymb,amsfonts,amsthm}
\usepackage{algorithmic}
\usepackage{graphicx}
\usepackage{textcomp}
\usepackage{xcolor}
\def\BibTeX{{\rm B\kern-.05em{\sc i\kern-.025em b}\kern-.08em
    T\kern-.1667em\lower.7ex\hbox{E}\kern-.125emX}}

\begin{document}

\title{On the explainability of max-plus neural networks
{
}
\thanks{This work was supported by Hi!Paris  and by PGMO (grant P-2025-0028)}
}

\author{\IEEEauthorblockN{Ikhlas Enaieh}
\IEEEauthorblockA{\textit{LTCI, Télécom Paris} \\
\textit{Institut Polytechnique de Paris}\\
Palaiseau, France \\
ikhlas.enaieh@telecom-paris.fr}
\and
\IEEEauthorblockN{Olivier Fercoq}
\IEEEauthorblockA{\textit{LTCI, Télécom Paris} \\
	\textit{Institut Polytechnique de Paris}\\
	Palaiseau, France \\
olivier.fercoq@telecom-paris.fr}
\and
\IEEEauthorblockN{Ángel García Pedrero}
\IEEEauthorblockA{\textit{DATSI} \\
	\textit{Universidad Politécnica de Madrid}\\
	Madrid, Spain \\
	angel.garcia@ctb.upm.es}
}
\maketitle

\begin{abstract}
We investigate the explanability properties of the recently proposed linear-min-max neural networks. At initialization, they can be interpreted as k-medoids with the infinity norm as a distance. Then, they are trained using subgradient descent to better fit the data. The model has been shown to be a universal approximator. Yet, we can trace the decision process because a single most activated neuron is responsible for the value of the output. Using this property, we designed a pixel fragility measure that determines whether changes to a single pixel may be responsible to a change in the classification output. Experiments on the PneumoniaMnist dataset show that this explanation for the output of the neural network compares favorably to SHAP and Integrated Gradient.
\end{abstract}

\begin{IEEEkeywords}
Max-plus neural network, explainable artificial intelligence, k-medoids
\end{IEEEkeywords}

\section{Introduction}

The remarkable success of deep learning in a wide range of applications has been accompanied by growing concerns about the interpretability and transparency of neural network decisions. As models become more complex, the need for explainable AI (XAI) has intensified, particularly in critical domains such as healthcare, finance, and autonomous systems. Explainable AI aims to provide human-understandable insights into model behavior, fostering trust and enabling domain experts to validate and refine model decisions \cite{molnar2020interpretable,guidotti2018survey}.
Our goal in this paper is to show that max-plus neural networks (also known as morphological neural networks)~\cite{ritter1996introduction} is a promising class of models that inherently support interpretability. These networks are built upon the mathematical framework of max-plus algebra, which replaces traditional addition and multiplication with the maximum and addition operations, respectively \cite{gaubert1997methods}. This structure not only preserves the universal approximation capabilities of conventional neural networks~\cite{minmaxplus} but also offers a unique form of interpretability: the decision process can often be traced back to a single, most activated neuron, making it easier to understand and visualize the model’s reasoning. Their training is challenging 
but \cite{enaieh2026exploiting} showed that using a well designed initialization and a sparse subgradient descent method, it is not out of reach, even for rather large datasets.

In this paper, we investigate the explainability properties of the recently proposed linear-min-max neural networks, a variant of max-plus networks. We first show that the initialization proposed in~\cite{enaieh2026exploiting} amounts to building a k-medoids clustering at initialization, with the infinity norm as a distance metric. After training, this interpretation is lost but we can still identify which neuron is responsible for the output. We also leverage their inherent interpretability to design a pixel fragility measure, which identifies whether changes to a single pixel can alter the classification output. Our experiments on the PneumoniaMnist dataset demonstrate that this approach provides explanations that compare favorably to established methods such as SHAP~\cite{lundberg2017unified} and Integrated Gradients~\cite{sundararajan2017axiomatic}.

\section{Linear-Min-Max-Plus neural networks}

Given an input vector $\mathbf{x} \in \mathbb{R}^{N}$, 
a weight vector $\mathbf{w} \in \mathbb{R}^{N}$, and a bias $b \in \mathbb{R}$, 
the morphological perceptron \cite{morphological_activation} computes its activation as
$
a(\mathbf{x}) = \max \left\{ b, \; \underset{1 \leq i \leq {\color{black}N}\ 
}{\max}  \{ x_i + w_i \} \right\}$,
where $x_i$ (resp.\ $w_i$) is the $i$-th component of $\mathbf{x}$ (resp.\ $\mathbf{w}$).

The Linear-Min-Max-Plus neural network defined in \cite{minmaxplus} for regression tasks and extended in \cite{enaieh2026exploiting} for classification is composed of three layers.

\begin{itemize}
	\item \textbf{Linear Layer:} Let ${\color{black}P}$ denote the number of input features and $x_n = X_{n,:}=(X_{n, 1}, \dots, X_{n,P})\in\mathbb{R}^{\color{black}P}$ the $n$-th sample.
	We apply a sparse linear transformation
	\[
	\lambda(x) = [\,K_{0}x_1,\,-K_{1}x_1,\,\dots,\,K_{2{\color{black}P}-1}x_{\color{black}P},\,-K_{2{\color{black}P}}x_{\color{black}P}\,],
	\]
	where $K$ is a vector of parameters. This can be also written as
	\[
	\lambda(x) = W^0 x = (W^{0,+}x, W^{0, -}x) \in \mathbb{R}^{2{\color{black}P}}, 
	 W^0 \in \mathbb{R}^{2P \times {\color{black}P}},
	\]
	where $W^0$ is a sparse matrix with predefined sparsity pattern.
	
	\item \textbf{Min-Plus Layer:} Let $ h = \{1,\dots,H_1\}$ denote the hidden neurons in this layer. Then :
	\[
	g_{h}(x) \;=\; \min_{i \in \{1,\dots,2{\color{black}P}\}} 
	\big( \lambda_i(x) + W^1_{i,h} \big),
	\]
	where $W^1 = (W^{1,+}, W^{1,-}) \in \mathbb{R}^{2{\color{black}P} \times H_1}$ is the (min,+) weight matrix.
	
\item \textbf{Max-Plus Layer \& Softmax:} For each class $d \in \{1, \dots, C\}$, the class score is
\[
{\color{black}z}_{d}(x) \;=\; \max_{h \in \{1,\dots,H_1\}} 
\big( g_{h}(x) + {\color{black}W}^2_{h,d} \big),
\]
with $W^2 \in \mathbb{R}^{H_1 \times C}$, and the predicted probabilities are
\[
\hat{{\color{black}y}}_{d}(x) \;=\; 
\frac{\exp\!\big( {\color{black}z}_{d}(x) \big)}
{\sum_{d'=1}^{C} \exp\!\big( {\color{black}z}_{d'}(x) \big)}.
\]

\end{itemize}

It has been shown in \cite{minmaxplus} that Linear-Min-Max-Plus networks can approximate any Lipschitz continuous function over a compact set. This universal approximation theorem shows that the model is rich enough to cover any regression or classification tasks and that with a sufficiently large number of hidden neurons, it can attain a high accuracy on the training set.

\section{Interpreting the initialization as K-medoids}

The loss of a Linear-Min-Max-Plus network is fundamentally non-convex and non-smooth with many local minimizers. Despite this challenging landscape, we can train the network using a sparse subgradient descent method~\cite{enaieh2026exploiting}, even on medium-size datasets. 

Yet, as shown in \cite{InfluenceofInitialization}, max-plus neural network training is very sensitive to the initialization. Enaieh and Fercoq proposed in \cite{enaieh2026exploiting} to initialize the weights using a subset $(\bar x^n)_{1 \leq n \leq H_1}$ of the samples together with their class $(\bar y^n)_{1 \leq n \leq H_1}$. They set $K_i = K$ for a given $K \in \mathbb R$, $W^1_{i,h} = -\lambda_i(\bar x^h)$ and $W^2_{h, \bar y^h} = K$,  $W^2_{h, d} = -K$ for $d \neq \bar y^h$. This initialization is motivated by the fact that when $H_1$ is equal to the size of the training set, then the universal approximation theorem ensures a perfect classification on the training set.

Here, we will show that this initialization amounts to defining the $(\bar x^n)$ as $K$-medoids \cite{rdusseeun1987clustering} with the infinity norm.
Indeed, at initialization, the output of the linear-min-max-plus network is given by
\begin{multline*}
z_d(x) = \max_{1\leq h\leq H_1} \min_{1\leq p\leq P} \min(K x_p -
K \bar x_p^h, K\bar x_p^h - Kx_p) + W^2_{h,d}\end{multline*}
By using the facts that $\min(x_1, x_2) = -\max(-x_1, -x_2)$ and that for $K$ large enough, $W^2_{h,d}$ takes much bigger values when $d = \bar y^h$ than in the other case, we obtain
\begin{align*}
z_d(x) &= -K\min_{h : \bar y^h = d} \max_{p \in \{1, \ldots, P\}} \max(-x_p +
\bar x_p^h, -\bar x_p^h +x_p) \\
& =- K \min_{h : \bar y^h = d} \|x - \bar x^h\|_\infty
\end{align*}
Thus at initialization, the linear-min-max-plus network performs $C$ $K$-medoid clusterings, one for each class. The predicted class is the one of the closest medoid. 

Then, the training is a refinement of these $K$-medoid clusters where we loose the $K$-medoid interpretation. However, what remains is that each min-plus neuron is associated to a specific class where its $W^2$ weight is larger and, moreover, the logits are completely determined by the value of the most activated neuron of the min-plus layer.

\section{Pixel fragility}

In this section, we introduce a pixel fragility measure, which is specific to linear-min-max-plus networks. It is based on the fact that those networks have a clear activation structure, where one single neuron of each layer conveys the value of the layer.

We define the sensitivity $S_{p, h}(x)$ of pixel $p$ for the image $x$ and neuron $h$ as
the amount that one needs to change the value $v$ of the pixel for the activation $g_h(\lambda(x + v \delta_p))$ to be different from $g_h(\lambda(x))$.

The value $v$ satisfies the inequalities:
\begin{align*}
W^{0, +}_p (x_{p} + v) + W^{1, +}_{h,p} \geq g_h(\lambda (x)) \\
W^{0, -}_p (x_{p} + v) + W^{1, -}_{h,p} \geq g_h(\lambda (x))
\end{align*}
Since $W^{0, -}_{p, h} < 0$ and $W^{0, +}_{p, h} > 0$, this is equivalent to
\begin{align*}
\frac{g_h(\lambda (x)) -  W^{1, +}_{h,p}}{W^{0, +}_p} - x_p \leq v \leq  \frac{g_h(\lambda (x)) -  W^{1, -}_{h,p}}{W^{0, -}_p} - x_p
\end{align*}
The sensitivity is then defined as
$S_{p, h}(x) = \min\Big(x_p - \frac{g_h(\lambda (x_{n, :})) -  W^{1, +}_{h,p}}{W^{0, +}_p} , \frac{W^{1, -}_{h,p} - g_h(\lambda (x_{n, :}))}{|W^{0, -}_p|} - x_{p}\Big)$


Let $c$ be the class predicted by the neural network for the image $x$ and $-c$ the other class. Since the min-plus neuron $h$ is typed by its largest value in $W^2_{h, :}$, we can define the sets $H_c$ and $H_{-c}$ of the neurons of each type. We also denote $d(h) = \arg\max_d W^2_{h,d}$.
We now define the extended sensitivity as the amount of change in a single pixel for the activation of its currently activated neuron to reach the largest activation $z_c(x)$.
Similarly to $S_{p, h}(x)$, its formula is given by
\begin{multline*}
s_{h,c}(x)  = z_c(x) - g_h(\lambda(x)) - W^2_{h, d(h)} \\
\bar S_{p, h}(x) = \min\Big( x_{p} - \frac{g_h(\lambda (x_{n, :})) - s_{h,c}(x) -  W^{1, +}_{h,p}}{W^{0, +}_p},\\ \frac{ s_{h,c}(x) + W^{1, -}_{h,p} - g_h(\lambda (x_{n, :}))}{|W^{0, -}_p|} - x_{p}\Big)
\end{multline*}
Note that $s_{h,c}(x) \geq 0$ and that $\bar S_{p,h}(x) \geq S_{p,h}(x)$.

Then we define the pixel fragility of pixel $p$ for image $x$ as
\begin{align*}
	F_p(x) = \min_{h \in H_{-c}} \bar S_{p,h}(x)
\end{align*}
This corresponds to how much the value of pixel $p$ should be changed for the max-neuron of the opposite class to activate at the same level as the current activated neuron.

\section{Numerical experiment}

\subsection{Dataset and models}

We consider the PneumoniaMnist dataset~\cite{yang2023medmnist}, which consists of labeled X-ray images of chests with 28x28 pixels each. There are 4708 images in the training set, 524 in the validation set and 624 images in the test set.

Using the training set, we trained two models:
\begin{itemize}
	\item a 2-layer perceptron (MLP) with 26 hidden neurons, 2 output neurons and rectified linear units activations using the Adam optimizer \cite{kingma2014adam}, with confusion matrices $\begin{bmatrix}1158 & 56 \\ 
    83& 3411 \end{bmatrix}$ on the training set and $\begin{bmatrix} 151 &  83 \\
         8 & 382 \end{bmatrix}$ on the testing set;
	\item a Linear-Min-Max model (LMM) with 25 min-plus neurons and 2 max-plus neurons using the sparse subgradient method of~\cite{enaieh2026exploiting}, with confusion matrices $\begin{bmatrix}1064 &  150 \\ 262 & 3232  \end{bmatrix}$ on the training set and $\begin{bmatrix} 131 & 103 \\ 
       34 & 356\end{bmatrix}$ on the testing set.
\end{itemize}
Both models have the same number of trainable parameters: 40,818. MLPs tend to be overconfident while LMMs tend to be underconfident. Thus, we calibrate the temperature of their softmax function \cite{guo2017calibration} in such a way that, in average, the prediction probability they output is 0.8.


\subsection{Explainable AI metrics}

We have compared our new pixel fragility measure with two other explainable AI techniques: SHAP~\cite{lundberg2017unified} and Integrated Gradients~\cite{sundararajan2017axiomatic}.
They apply to both MLP and LMM because they are model-agnostic and do not rely on structural assumption on the network like gradCAM~\cite{selvaraju2016grad} does.
For a quantitative comparison, we used the metrics introduced in~\cite{skliarov2025comparative}. We can see that for all of them, stability is between 1 and 2. This means that for images that differ slightly, so do their explanations. The most informative metric is fidelity. To compute it, we first rank the pixels from the most informative to the least informative according to the explanation. Then, we gradually set the pixels to a default value (here we chose 0.5, i.e. gray) and recompute the predictions. Fidelity is then average prediction probability over samples and pixel deletions. A fidelity score close to 0.8 means that the explanation is meaningless. Indeed, without the ``important'' pixels, the output is the same. On the other hand, a fidelity close to 0.5 means that the network is unable to classify samples where ``important'' pixels have been removed.

We computed the metrics using the test dataset and adapted the code from~\cite{skliarov2025comparative}.
We can see on Table~\ref{tab:xaimetrics4mlp} that SHAP and IntGrad both work well for explaining the outcome of the MLP. They really show the pixels that are used for classification. 

The results for LMM are reported on Table~\ref{tab:xaimetrics4lmm}, IntGrad suffers from the fact that subgradient are sparse. Indeed, as can be seen on Figure~\ref{fig:lmm}, its explanation focus on too few pixels and this is not enough to be significant in the fidelity score. SHAP is completely model-agnostic and works well also for LMM. 
We can see that our model-specific pixel fragility measure is using the structure of the network in an efficient manner. Its fidelity score is even closer to 0.5 than SHAP's, with a computational cost which is much smaller.

\begin{table}[htbp]
	\begin{tabular}{lrrr}
	                & Stability & Fidelity & Time (s/image)\\
		SHAP on LMM     & 2.64      & 0.504    & 0.110 \\
		IntGrad on LMM  & 1.51      & 0.402    & 0.005 \\
	\end{tabular}
	\caption{XAI metrics for MLP}
	\label{tab:xaimetrics4mlp}
\end{table}

\begin{table}[htbp]
	\begin{tabular}{lrrr}
		              & Stability & Fidelity & Time (s/image)\\
		SHAP on LMM    & 2.04      & 0.559    & 0.220 \\
		IntGrad on LMM  & 1.63      & 0.736    & 0.010 \\
		Pixel Fragility on LMM& 1.64      & \bf{0.540}    & 0.005 \\
	\end{tabular}
\caption{XAI metrics for LMM}
\label{tab:xaimetrics4lmm}
\end{table}

\subsection{Examples of explanations}

We show here the explanations for 4 samples: sample 0 (class 1, predicted 1), sample 1 (class 0, predicted 1), sample 3 (class 0, predicted 0), sample 176 (class 1, predicted 0). We chose them because both MLP and LMM assigned the same prediction for those samples.

\begin{figure}
\includegraphics[width=0.9\columnwidth]{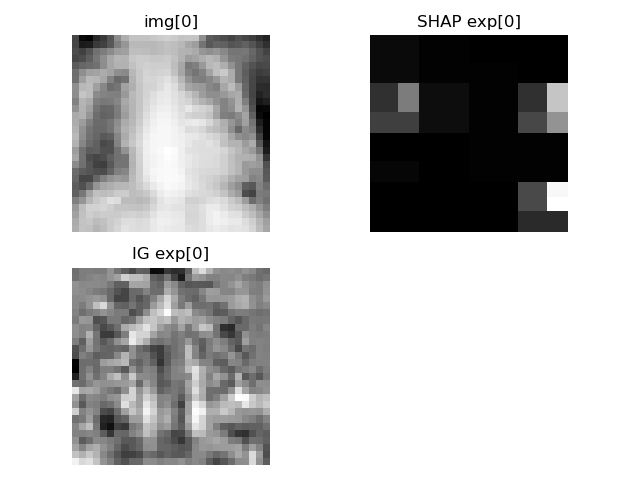}
\includegraphics[width=0.9\columnwidth]{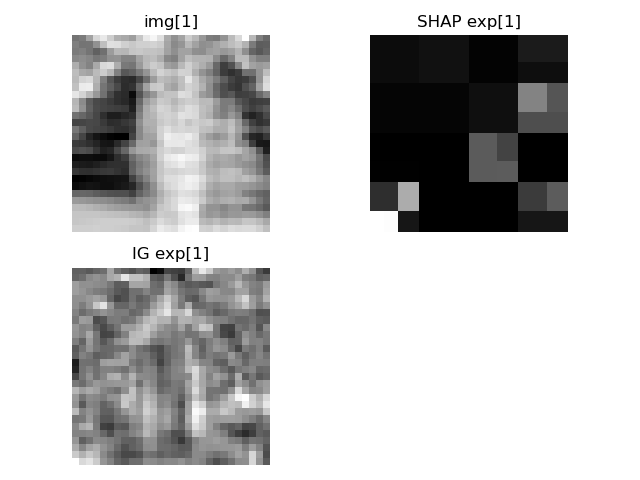}
\includegraphics[width=0.9\columnwidth]{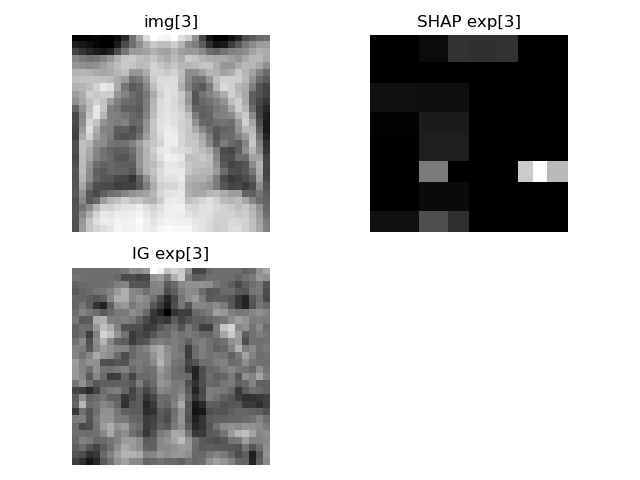}
\includegraphics[width=0.9\columnwidth]{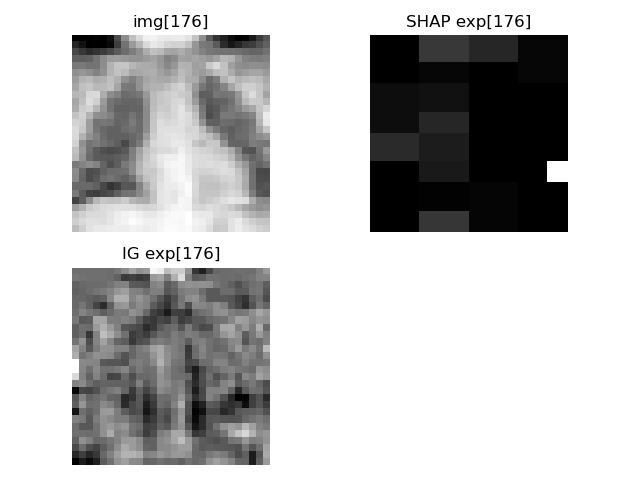}
\caption{Examples of SHAP and IntGrad explanations for the multi-layer perceptron (lighter means more important)} 
\label{fig:nlp}
\end{figure}

In Figure~\ref{fig:nlp}, we present the importance maps for MLP.
We remark that IntGrad importance maps are very noisy and difficult to interpret. SHAP importance maps put a very strong emphasis on particular positions, sometimes on the boundary of the image.

\begin{figure}
	\includegraphics[width=0.9\columnwidth]{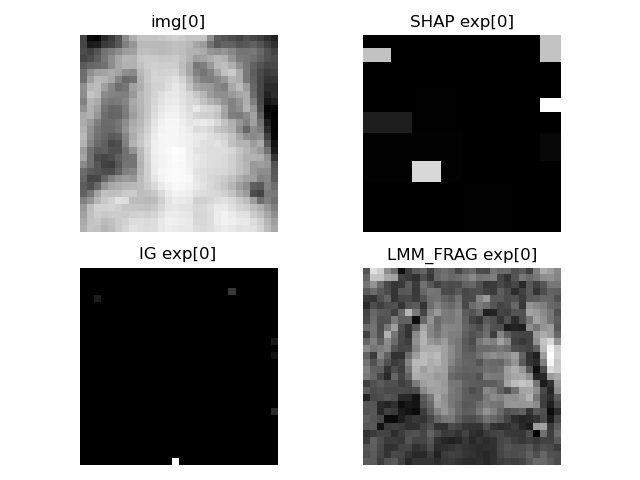}
	\includegraphics[width=0.9\columnwidth]{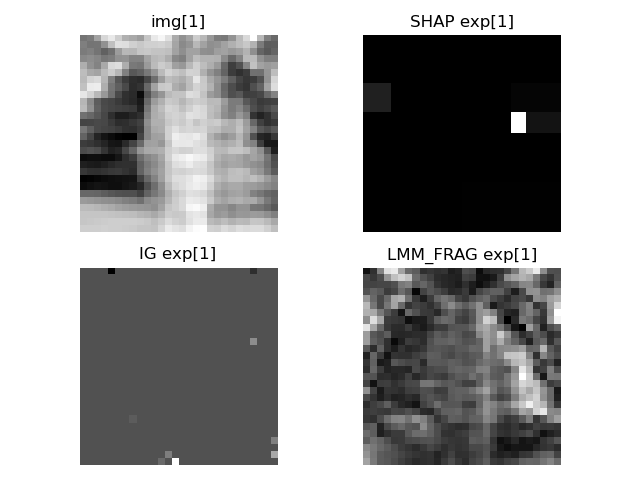}
	\includegraphics[width=0.9\columnwidth]{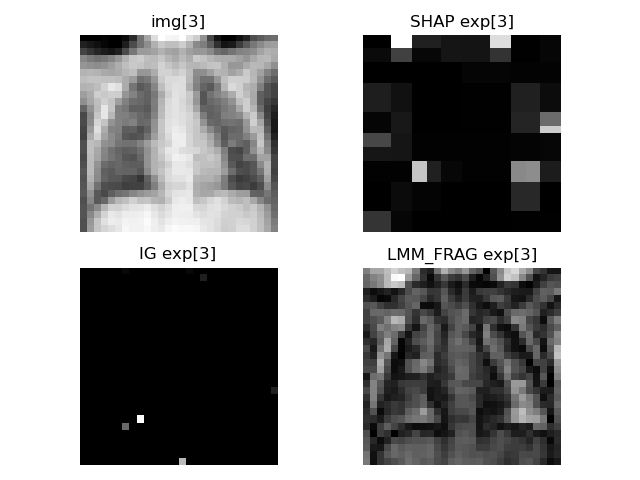}
	\includegraphics[width=0.9\columnwidth]{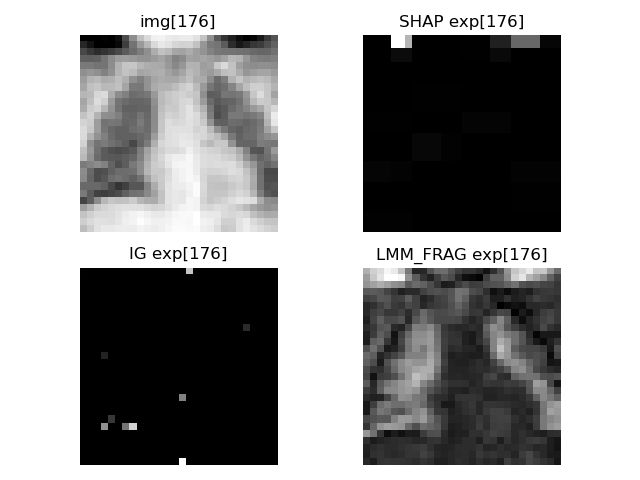}
\caption{Examples of SHAP, IntGrad and Pixel Fragility explanations for the linear-min-max-plus network (lighter means more important)} 
\label{fig:lmm}
\end{figure}

On Figure~\ref{fig:lmm}, we present examples of explanations for output of the Linear-Min-Max-Plus networks. Note that the SHAP and IntGrad importance maps are not the same as for the MLP. This is a sign that both models operate in a different manner and focus on different parts of the images to make the predictions.

The LMM Pixel Fragility importance maps seem a promising alternative. We can see that even though some artifact remain, like a strong focus on the background for samples 3 and 176, lungs are also a clear focus for all the samples.

\bibliographystyle{plain}
\bibliography{literature}

\begin{thebibliography}{10}

\bibitem{InfluenceofInitialization}
Mihaela Dimitrova, Samy Blusseau, and Santiago Velasco-Forero.
\newblock Learning morphological representations of image transformations:
  Influence of initialization and layer differentiability.
\newblock In {\em International Conference on Discrete Geometry and
  Mathematical Morphology}, pages 371--383, 2025.

\bibitem{enaieh2026exploiting}
Ikhlas Enaieh and Olivier Fercoq.
\newblock Exploiting subgradient sparsity in max-plus neural networks.
\newblock {\em hal-05502128 preprint}, 2026.

\bibitem{gaubert1997methods}
St{\'e}phane Gaubert and Max Plus.
\newblock Methods and applications of (max,+) linear algebra.
\newblock In {\em Annual symposium on theoretical aspects of computer science},
  pages 261--282. Springer, 1997.

\bibitem{guidotti2018survey}
Riccardo Guidotti, Anna Monreale, Salvatore Ruggieri, Franco Turini, Fosca
  Giannotti, and Dino Pedreschi.
\newblock A survey of methods for explaining black box models.
\newblock {\em ACM computing surveys (CSUR)}, 51(5):1--42, 2018.

\bibitem{guo2017calibration}
Chuan Guo, Geoff Pleiss, Yu~Sun, and Kilian~Q Weinberger.
\newblock On calibration of modern neural networks.
\newblock In {\em International conference on machine learning}, pages
  1321--1330. PMLR, 2017.

\bibitem{kingma2014adam}
Diederik~P Kingma and Jimmy Ba.
\newblock Adam: A method for stochastic optimization.
\newblock In {\em Proc. of ICLR}, 2015.

\bibitem{lundberg2017unified}
Scott~M Lundberg and Su-In Lee.
\newblock A unified approach to interpreting model predictions.
\newblock {\em Advances in neural information processing systems}, 30, 2017.

\bibitem{minmaxplus}
Ye~Luo and Shiqing Fan.
\newblock Min-max-plus neural networks.
\newblock {\em preprint arXiv:2102.06358}, 2021.

\bibitem{molnar2020interpretable}
Christoph Molnar.
\newblock {\em Interpretable machine learning}.
\newblock Lulu. com, 2020.

\bibitem{morphological_activation}
Ranjan Mondal, Soumendu~Sundar Mukherjee, Sanchayan Santra, and Bhabatosh
  Chanda.
\newblock Morphological network: How far can we go with morphological neurons?
\newblock In {\em British Machine Vision Conference}, 2019.

\bibitem{rdusseeun1987clustering}
LKPJ Rdusseeun and P~Kaufman.
\newblock Clustering by means of medoids.
\newblock In {\em Proceedings of the statistical data analysis based on the L1
  norm conference, neuchatel, switzerland}, volume~31, page~28, 1987.

\bibitem{ritter1996introduction}
Gerhard~X Ritter and Peter Sussner.
\newblock An introduction to morphological neural networks.
\newblock In {\em Proceedings of 13th International Conference on Pattern
  Recognition}, volume~4, pages 709--717. IEEE, 1996.

\bibitem{selvaraju2016grad}
Ramprasaath~R Selvaraju, Abhishek Das, Ramakrishna Vedantam, Michael Cogswell,
  Devi Parikh, and Dhruv Batra.
\newblock Grad-cam: Why did you say that?
\newblock {\em arXiv preprint arXiv:1611.07450}, 2016.

\bibitem{skliarov2025comparative}
Mykyta Skliarov, Radwa~El Shawi, Chedia Dhaoui, and Nada Ahmed.
\newblock A comparative evaluation of explainability techniques for image data.
\newblock {\em Scientific Reports}, 15(1):41898, 2025.

\bibitem{sundararajan2017axiomatic}
Mukund Sundararajan, Ankur Taly, and Qiqi Yan.
\newblock Axiomatic attribution for deep networks.
\newblock In {\em International conference on machine learning}, pages
  3319--3328. PMLR, 2017.

\bibitem{yang2023medmnist}
Jiancheng Yang, Rui Shi, Donglai Wei, Zequan Liu, Lin Zhao, Bilian Ke,
  Hanspeter Pfister, and Bingbing Ni.
\newblock Medmnist v2-a large-scale lightweight benchmark for 2d and 3d
  biomedical image classification.
\newblock {\em Scientific data}, 10(1):41, 2023.

\end{thebibliography}

\end{document}